# AdaBatch: Adaptive Batch Sizes for Training Deep Neural Networks

**Aditya Devarakonda**
Computer Science Division
University of California, Berkeley
aditya@cs.berkeley.edu

**Maxim Naumov & Michael Garland**
NVIDIA
Santa Clara, CA 95050, USA
{mnaumov, mgarland}@nvidia.com

## Abstract

Training deep neural networks with Stochastic Gradient Descent, or its variants, requires careful choice of both learning rate and batch size. While smaller batch sizes generally converge in fewer training epochs, larger batch sizes offer more parallelism and hence better computational efficiency. We have developed a new training approach that, rather than statically choosing a single batch size for all epochs, adaptively increases the batch size during the training process. Our method delivers the convergence rate of small batch sizes while achieving performance similar to large batch sizes. We analyse our approach using the standard AlexNet, ResNet, and VGG networks operating on the popular CIFAR-10, CIFAR-100, and ImageNet datasets. Our results demonstrate that learning with adaptive batch sizes can improve performance by factors of up to 6.25 on 4 NVIDIA Tesla P100 GPUs while changing accuracy by less than 1% relative to training with fixed batch sizes.

## 1 Introduction

Stochastic Gradient Descent (SGD) and its variants are the most widely used techniques for training deep neural networks. Deep networks are large by design, and the training process requires large amounts of data. Therefore, implementations typically divide the training set into a (potentially large) series of batches of some fixed size. Each batch is processed in sequence during one training epoch; however, the individual training samples within a single batch may be processed in parallel (Bottou et al., 2016; Goodfellow et al., 2016).

The user overseeing the training process typically chooses a static batch size $r$, which is held constant throughout the training process. However, static batch sizes force the user to resolve an important conflict. On one hand, small batch sizes are desirable since they tend to produce convergence in fewer epochs (Das et al., 2016; Keskar et al., 2016). On the other hand, large batch sizes offer more data-parallelism which in turn improves computational efficiency and scalability (Goyal et al., 2017; You et al., 2017).

Our approach to resolving this trade-off between small and large batch sizes is to *adaptively* increase the batch size during training. We begin with a small batch size $r$, chosen to encourage rapid convergence in early epochs, and then progressively increase the batch size between selected epochs as training proceeds. For the experiments reported in this paper, we double the batch size at specific intervals and simultaneously adapt the learning rate $\alpha$ so that the ratio $\alpha/r$ remains constant. Our adaptive batch size technique has several advantages. It delivers the accuracy of training with small batch sizes, while improving performance by increasing the amount of work available per processor in later epochs. Furthermore, the large batches used in later epochs expose sufficient parallelism to create the opportunity for distributing work across many processors, where those are available. Our approach can also be combined with other existing techniques for constructing learning rate decay schedules to increase batch sizes even further.

We have applied our adaptive batch size method to training standard AlexNet, ResNet, and VGG networks on the CIFAR-10, CIFAR-100, and ImageNet datasets. Our experimental results, detailed in Section 4, demonstrate that training with adaptive batch sizes attains similar test accuracies with faster running times compared with training using fixed batch sizes. Furthermore, we experimentally show that adaptive batch sizes can be combined with other large batch size techniques to yield speedups of up to $6.25\times$ while leaving test accuracies relatively unchanged.

## 2 RELATED WORK

Previous work introduced gradual learning rate warmup and linear learning rate scaling in order to attain batch sizes of 8192 for ImageNet CNN training in a large, distributed GPU cluster setting (Goyal et al., 2017). More recently, the use of a layer-wise learning rate scaled by the norms of the gradients allowed even higher batch sizes of 32,768 (You et al., 2017). Both results use a fixed batch size throughout training whereas our work changes the batch sizes during training. Furthermore, we show that our work is complementary to these existing results.

The relationship between adaptive batch sizes and learning rates is well-known. In particular, this relationship was illustrated for strongly-convex, unconstrained optimization problems by Friedlander & Schmidt (2012). This work showed that batch size increases can be used instead of learning rate decreases. On the other hand, a batch size selection criterion based on estimates of gradient variance, which are used to adaptively increase batch sizes, was introduced by Byrd *et al.* (2012). Both approaches consider second-order, Newton-type methods. Our work complements and adds to this research by exploring adaptive batch sizes for various neural network architectures.

Several authors have also studied adaptively increasing batch size with fixed schedules from the context of accelerating the optimization method through variance reduction (Daneshmand et al., 2016; Harikandeh et al., 2015). Both works study the theoretical and empirical convergence behavior of their adaptive batch size optimization methods on convex optimization problems. In contrast an adaptive criterion to control the batch size increases and illustration of their convergence on convex problems and convolutional neural networks is developed by De *et al.* (2016) and Balles *et al.* (2017). While both illustrate the practicality of coupling adaptive batch sizes with learning rates, they do not explore the performance benefits that can be gained through the use of adaptive batch sizes when coupled with existing large batch size CNN training work (Goyal et al., 2017; You et al., 2017).

A very recent work illustrates that learning rate decay can be replaced with batch size increase (Smith et al., 2017). The aforementioned study shows that the batch size increase in lieu of learning rate decay works on several optimization algorithms: SGD, SGD with momentum, and Adam. Furthermore, it experiments with altering the momentum term with batch size increases and explores the effects on convergence.

In addition, our research explores the performance tradeoffs from using an adaptive batch size technique on popular CNNs and illustrates that our technique is complementary to existing fixed, large batch size training techniques. Our results also independently verify that adaptive batch size can practically replace learning rate decay and lead to performance improvements. In particular, we show that adaptive batch size schedules can yield speedups that learning rate schedules alone cannot achieve. Finally, by combining adaptive batch sizes with large batch size techniques we show that even larger speedups can be achieved with similar test error performance.

## 3 ADAPTIVE BATCH SIZING AND ITS EFFECTS

The batch size and learning rate are intimately related tuning parameters (You et al., 2017; Goyal et al., 2017). Recent work has shown that adapting the learning rate either layerwise or over several iterations can enable training with large batch sizes without sacrificing accuracy. We will illustrate that the relationship between batch size and learning rate extends even further to learning rate decay. The following analysis provides the basis for our adaptive batch size technique.

### 3.1 LEARNING RATE

In supervised learning, the training of neural networks consists of repeated forward and backward propagation passes over a labelled data set. The data set is often partitioned into training, validation and test parts, where the first is used for fitting the neural network function to the data and the others for verification of the results.

Let a training data set be composed of data samples $\{(\mathbf{x}, \mathbf{z}^*)\}$, which are pairs of known inputs $\mathbf{x} \in \mathbb{R}^n$ and outputs $\mathbf{z}^* \in \mathbb{R}^m$. Further, let these pairs be ordered and partitioned into $q$ disjoint batches of size $r$. For simplicity of presentation, we assume that the number of pairs is $qr$. In cases where $r$ does not divide the number of pairs evenly, implementations must in practice either pad the last batch or correctly handle truncated batches.

Training a neural network with weights $W$ can be interpreted as solving an optimization problem

$$\arg \min_W \mathcal{L} \qquad (1)$$

for some choice of loss function $\mathcal{L}$. This optimization problem is often solved with a stochastic gradient descent algorithm, where the weight updates at $i$-th iteration are performed using the following rule

$$W_{i+1} = W_i - \frac{\alpha}{r} \Delta W_i \qquad (2)$$

for an update matrix $\Delta W_i$ computed with batch-size $r$ and learning rate $\alpha$.

Notice that there is a clear relationship between the batch size and the learning rate. According to Equation (2) after $q$ iterations (i.e., one epoch) with a learning rate $\alpha$ and batch size $r$ we have

$$W_{i+q} = W_i - \frac{\alpha}{r} \sum_{j=1}^{q} \Delta W_{i+j} \qquad (3)$$

Suppose that we instead train with larger batch sizes by grouping $\beta > 1$ batches. Note that this results in an effective batch size of $\beta r$ and results in $\tilde{q} = q/\beta$ iterations for one epoch of training. Under this setting we can write

$$W_{i+\tilde{q}} = W_i - \frac{\tilde{\alpha}}{\beta r} \sum_{j=1}^{\tilde{q}} \Delta \widetilde{W}_j \qquad (4)$$

for an update matrix $\widetilde{W}_j$ computed with batch size $\beta r$ and learning rate $\tilde{\alpha}$. Notice that this can be re-written as an accumulation of $\beta$ gradients with batch size $r$ as follows:

$$W_{i+\tilde{q}} = W_i - \frac{\tilde{\alpha}}{\beta r} \sum_{j=1}^{\tilde{q}} \left( \sum_{k=1}^{\beta} \Delta W_{i'} \right) \qquad (5)$$

where $\widetilde{W}_j = \sum_{k=1}^{\beta} \Delta W_{i'}$ and index $i' = (j-1)\beta + k$. Notice that $W_{i+q}$ might be similar to $W_{i+\tilde{q}}$ only if we set the learning rate $\alpha = \tilde{\alpha}/\beta$ and assume that updates $\Delta W_i \approx \Delta W_{i'}$ are similar in both cases. This assumption was empirically shown to hold for fixed large batch size training with gradual learning rate warmup (Goyal et al., 2017) after the first few epochs of training.

Notice that the factor $1/\beta$ can be interpreted as a learning rate decay, when comparing Equation (3) and (5). This relationship has been used to justify linearly scaling the learning rate for large batch size training (Goyal et al., 2017). On the other hand, we take advantage of this relationship to illustrate that increasing the batch size can mimic learning rate decay. Naturally, the two approaches can be combined as we will show in Section 4. In our experiments, we will consider adaptive batch sizes which increase according to a fixed schedule. We ensure that the effective learning rates for fixed batch size vs. adaptive batch size experiments are fixed throughout the training process for fair comparison (see Section 4 for details).

### 3.2 TEST ACCURACY AND PERFORMANCE

Training with large batch sizes is attractive due to its performance benefits on modern deep learning hardware. For example, on GPUs larger batch sizes allow us to better utilize GPU memory bandwidth and improve computational throughput (NVIDIA, 2016; 2017). Larger batch sizes are especially important when distributing training across multiple GPUs or even multiple nodes since they can hide communication cost more effectively than small batch sizes.

However, the performance benefits of large batch sizes come at the cost of lower test accuracies since large batches tend to converge to sharper minima (Das et al., 2016; Keskar et al., 2016). Through the use of learning rate scaling (Goyal et al., 2017) and layer-wise adaptive learning rates (You et al., 2017), larger batch sizes can attain better accuracies. While both approaches increase the batch size, the batch size they use remains fixed throughout training. We propose an approach that adaptively changes the batch size and progressively exposes more parallelism. This approach also allows one to progressively add GPUs, if available, to the training process.

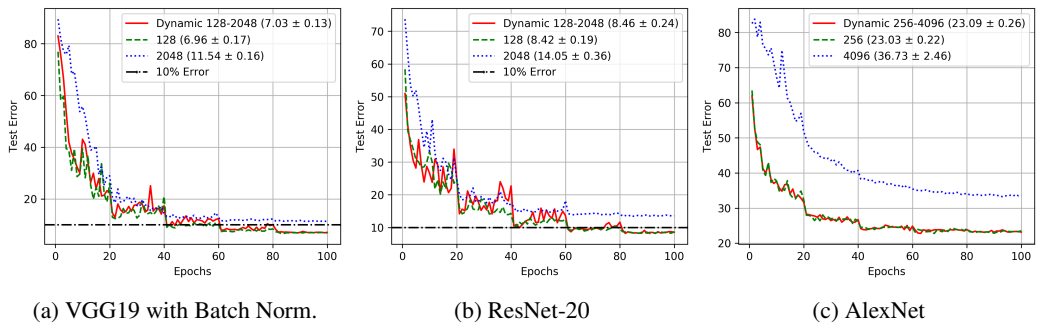

(a) VGG19 with Batch Norm.   (b) ResNet-20   (c) AlexNet

Figure 1: Comparison of CIFAR-10 test errors for adaptive versus fixed small and large batch sizes. The plots show the lowest test error and report mean ± standard deviation over 5 trials.

### 3.3 WORK PER EPOCH

Fixed batch sizes (small or large) require a fixed number of floating point operations (flops) per iteration throughout the training process. Since our technique adaptively increases the batch size, the flops per iteration progressively increases. Despite this increase, we can show that the flops per *epoch* remains fixed as long as the computation required for forward and backward propagation is a linear function of the batch size $r$.

For example, let us briefly illustrate this point on a fully connected layer with a weight matrix $W \in \mathbb{R}^{m \times n}$, input $X = [\mathbf{x}_1, ..., \mathbf{x}_r] \in \mathbb{R}^{n \times r}$ and error gradient $V = [\mathbf{v}_1, ..., \mathbf{v}_r] \in \mathbb{R}^{m \times r}$. Notice that for a batch of size $r$ the most computationally expensive operations during training are matrix-matrix multiplications

$$Y = WX \quad \text{and} \quad (6)$$
$$U = W^T V \quad (7)$$

in forward and backward propagation, respectively. These operations require $O(mnr)$ flops per iteration and $O(mnrq)$ flops per epoch. Notice that the amount of computation depends linearly on the batch size $r$. If we select a new, larger batch size of $\beta r$, then the flops per iteration increase to $O(mn\beta r)$. However, increasing the batch size by a factor of $\beta$ also reduces the number of iterations required by a factor of $\beta$. As a result the flops per epoch remains fixed at $O(mnrq)$ despite requiring more flops per iteration. Since larger batch sizes do not change the flops per epoch, they are likely to result in performance *improvements* due to better hardware efficiency.

A more detailed analysis of fully connected, convolutional and batch normalization layers can be found in the Appendix A, where we verify that the computation required for forward and backward propagation is indeed a linear function of the batch size. Our experimental results in Section 4 confirm these conclusions for CNNs.

### 4 EXPERIMENTAL RESULTS

In this section, we illustrate the accuracy tradeoffs and performance improvements of our adaptive batch size technique. We test this technique using the VGG (Simonyan & Zisserman, 2014), ResNet (He et al., 2016), and AlexNet (Krizhevsky et al., 2012) deep learning networks on CIFAR-10 (Krizhevsky et al., 2009a), CIFAR-100 (Krizhevsky et al., 2009b), and ImageNet (Deng et al., 2009) datasets. We implement our algorithms using PyTorch version 0.1.12 with GPU acceleration. Our experimental platform consists of 4 NVIDIA Tesla P100 GPUs interconnected via NVIDIA NVLink.

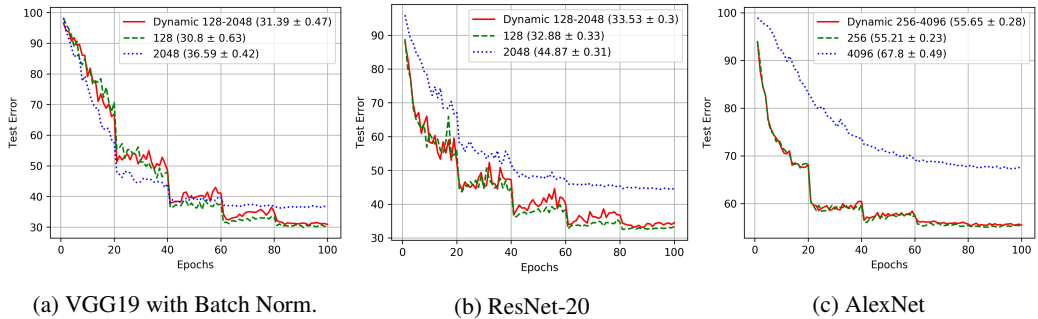

(a) VGG19 with Batch Norm.  (b) ResNet-20  (c) AlexNet

Figure 2: Comparison of CIFAR-100 test errors for adaptive versus fixed small and large batch sizes. The plots show the lowest test error and report mean $\pm$ standard deviation over 5 trials.

### 4.1 Fixed vs. Dynamic Batch Sizes

As we have illustrated in Section 2, learning rate decay is a widely used technique to avoid stagnation during training. While learning rate schedules may help improve test error, they rarely lead to faster training times. We begin by performing experiments to validate our claim that adaptive batch sizes can be used without significantly affecting test accuracy. For these experiments, we use SGD with momentum of $0.9$, weight decay of $5 \times 10^{-4}$, and perform 100 epochs of training. We use a base learning rate of $\alpha = 0.01$ and decay it every 20 epochs. For the adaptive method we decay the learning rate by $0.75$ and simultaneously double the batch size at the same 20-epoch intervals. The learning rate decay of $0.75$ and batch size doubling combine for an *effective* learning rate decay of $0.375$; therefore, we use a learning rate decay of $0.375$ for the fixed batch size experiments for the most direct comparison. All experiments in this section are performed on a single Tesla P100.

Figure 1 shows the test error on the CIFAR-10 dataset for (1a) VGG19 with batch normalization, (1b) ResNet-20, and (1c) AlexNet. For AlexNet the fixed batch sizes are 512 and 8192. For VGG19 and ResNet-20 the fixed batch sizes are 256 and 4096, which are smaller due to the constraint of fitting within the memory of a single GPU. Figure 1 plots the best test error for each batch size setting, but reports the mean and standard deviation over five trials in the legends. The noticeable drops in test error every 20 epochs are due to the learning rate decay. We observed that the adaptive batch size technique attained mean test errors within 1% of the smallest fixed batch size. Compared to the largest fixed batch sizes, the adaptive technique attained significantly lower test errors. Note that the fixed batch size experiments for VGG19 and ResNet-20 attain test errors comparable to those reported in prior work (He et al., 2016).

Figure 2 shows similar results on the CIFAR-100 dataset for the same networks and batch size settings. Once again, we see that the adaptive batch size technique attains test errors within 1% of the smallest fixed batch size. Our results indicate that learning rate schedules and batch size schedules are related and complementary. Both can be used to achieve similar effects on test accuracy. However, adapting the batch size provides the additional advantage of better efficiency and scalability without the need to sacrifice test error.

| Network | Batch Size | Forward Time (speedup) | Backward Time (speedup) |
|---|---|---|---|
| VGG19_BN | 128 | 933.79 sec. (1×) | 1571.35 sec. (1×) |
| | 128-2048 | **707.13 sec.** (1.32×) | **1322.59 sec.** (1.19×) |
| ResNet-20 | 128 | 256.59 sec. (1×) | 661.35 sec. (1×) |
| | 128-2048 | **218.97 sec.** (1.17×) | **578.63 sec.** (1.14×) |
| AlexNet | 256 | 66.24 sec. (1×) | 129.39 sec. (1×) |
| | 256-4096 | **44.34 sec.** (1.49×) | **89.69 sec.** (1.44×) |

Table 1: Comparison of CIFAR-100 forward and backward propagation running time over 100 epochs for adaptive versus fixed batch sizes. The table shows mean over 5 trials.

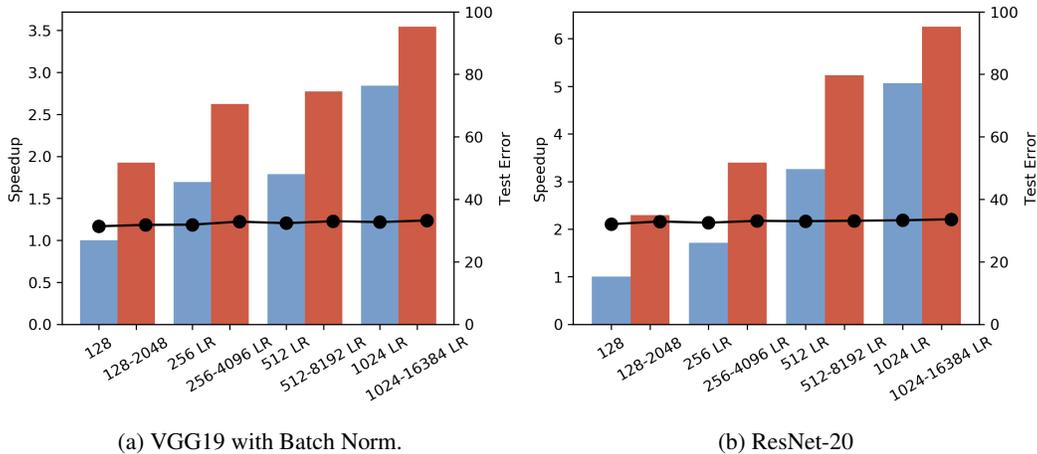

(a) VGG19 with Batch Norm.

(b) ResNet-20

Figure 3: Comparison of CIFAR-100 speedup (left vertical axis) and test errors (right vertical axis) for adaptive (in red) vs. fixed batch sizes (in blue), where "LR" uses gradual learning rate scaling for the first 5 epochs.

Table 1 quantifies the efficiency improvements that come from adaptive batch sizes. We report the running times on the CIFAR-100 dataset over 100 epochs of training. We omit the largest batch sizes from the table since they do not achieve comparable test errors. We also omit CIFAR-10 performance results since it is the same dataset. For all networks tested, we observed that the mean forward and backward propagation running times of adaptive batch sizes were better.

### 4.2 MULTI-GPU PERFORMANCE

While the speedups on a single GPU are modest, the ability to use batch sizes up to 4096 allows for better scalability in multi-GPU settings. As we have illustrated, adaptively increasing the batch size can act like a learning rate decay. In particular, we have experimentally shown that doubling the batch size behaves similarly to halving the learning rate. This suggests that our adaptive batch size technique can be applied to existing large batch size training techniques (Goyal et al., 2017; You et al., 2017). In this section, we will combine the former approach with our adaptive batch size technique and explore the test error and performance tradeoffs. Note that we do not use the latter approach since we would like to ensure our distributed batch sizes fit in each GPU's memory. We use PyTorch's `torch.nn.DataParallel` facility to parallelize across the 4 Tesla P100 GPUs in our test system.

We perform our experiments on CIFAR-100 using VGG19 with Batch Normalization and ResNet-20. We use SGD with momentum of $0.9$ and weight decay of $5 \times 10^{-4}$. The baseline settings for both networks are fixed batch sizes of $128$, base learning rate of $0.1$, and learning rate decay by a factor of $0.25$ every $20$ epochs. The adaptive batch size experiments start with large initial batch sizes, perform gradual learning rate scaling over $5$ epochs and double the batch every $20$ epochs and decay learning rate by $0.5$. We perform $100$ epochs of training for all settings.

Figure 3 shows the speedups (left vertical axis) and test errors (right vertical axis) on (3a) VGG19 and (3b) ResNet-20. All speedups are normalized against the baseline fixed batch size of $128$. The additional "LR" labels on the horizontal axis indicate settings which require a gradual learning rate scaling in the first $5$ epochs. Compared to the baseline fixed batch size setting, we see that adaptive 1024–16384 batch size attains average speedups (over 5 trials) of $3.54\times$ (VGG19) and $6.25\times$ (ResNet-20) with less than $2\%$ difference in test error. Note that the speedups are attained due to the well-known observation that large batch sizes train faster (Goyal et al., 2017; Keskar et al., 2016; Das et al., 2016). It is also useful to note that our approach can scale to more GPUs due to the use of progressively large batch sizes.

Figure 4 shows the test error curves for $4$ batch size settings: fixed 128, adaptive 128–2048, fixed 1024 with learning rate warmup, and adaptive 1024–16384 with learning rate warmup. We report results for VGG19 with batch norm and ResNet-20 on the CIFAR-100 dataset. These experiments illustrate that adaptive batch sizes converge to test errors that are similar ($< 1\%$ difference) to their fixed batch size counterparts. Furthermore, the results indicate that adaptive batch sizes can be coupled with the gradual learning rate warmup technique (Goyal et al., 2017) to yield progressively larger batch sizes during training. We conjecture that our adaptive batch size technique can also be coupled with layer-wise learning rates (You et al., 2017) for even larger batch size training.

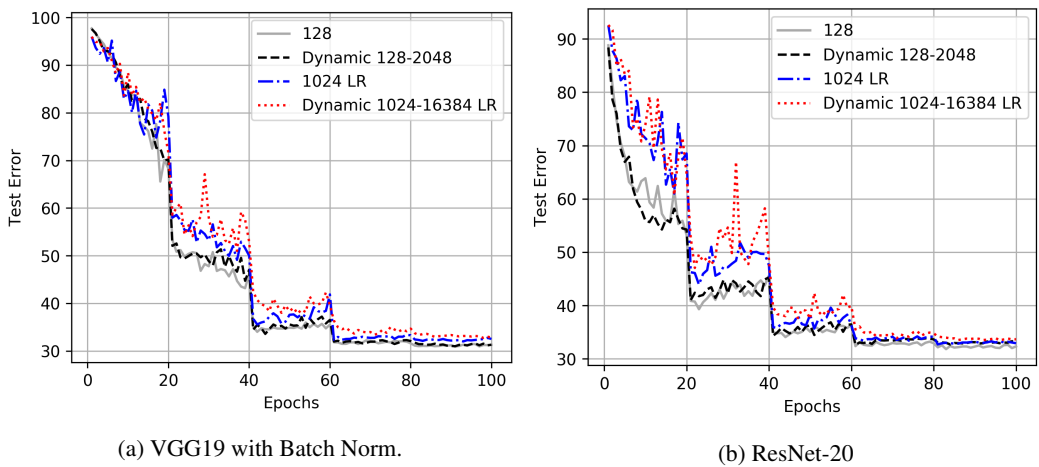

(a) VGG19 with Batch Norm.

(b) ResNet-20

Figure 4: Comparison of CIFAR-100 test errors curves for adaptive versus fixed batch sizes.

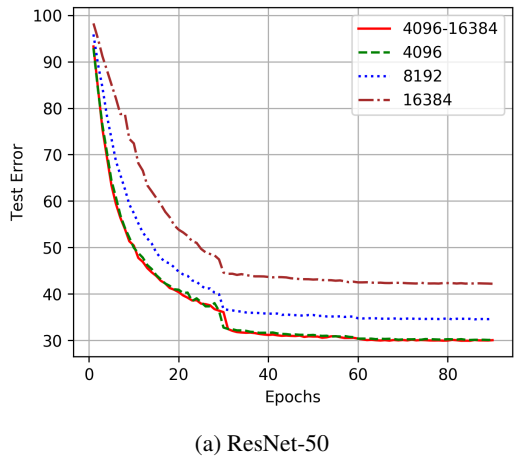

(a) ResNet-50

Figure 5: Comparison of ImageNet test errors curves for adaptive versus fixed batch sizes.

## 4.3 IMAGENET TRAINING WITH ADABATCH

In this section, we illustrate the accuracy and convergence of AdaBatch on ImageNet training with the ResNet-50 network. Once again we use PyTorch as the deep learning framework and its `DataParallel` API to parallelize over 4 NVIDIA Tesla P100 GPUs. Due to the large number of parameters, we are only able to fit a batch of $512$ in multi-GPU memory. When training batch sizes $> 512$ we choose to accumulate gradients. For example, when training with a batch size of $1024$ we perform two forward and backward passes with batch size $512$ and accumulate the gradients before updating the weights. The effective batch size for training is thus $1024$, as desired, however the computations are split into two iterations with batch size $512$. Due to this gradient accumulation we do not report performance results.

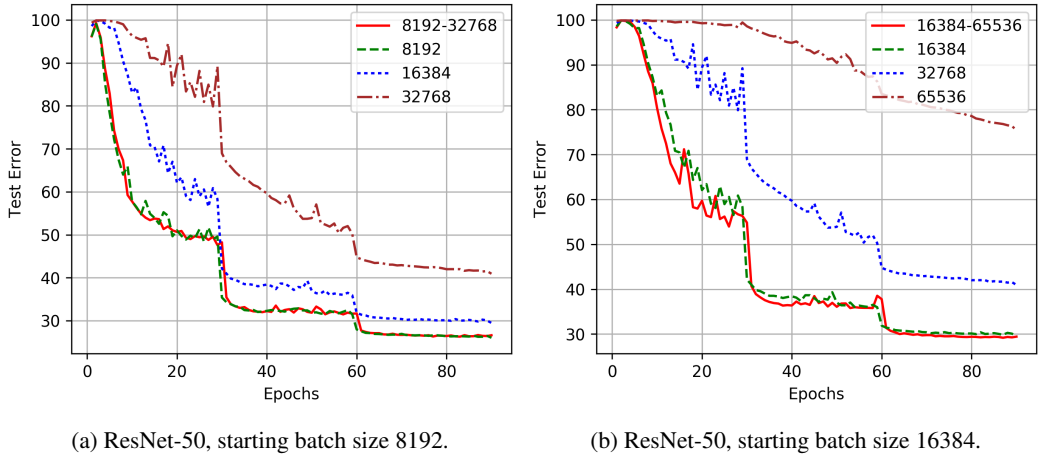

(a) ResNet-50, starting batch size 8192.

(b) ResNet-50, starting batch size 16384.

Figure 6: Comparison of ImageNet test errors curves for adaptive versus fixed batch sizes with LR warmup.

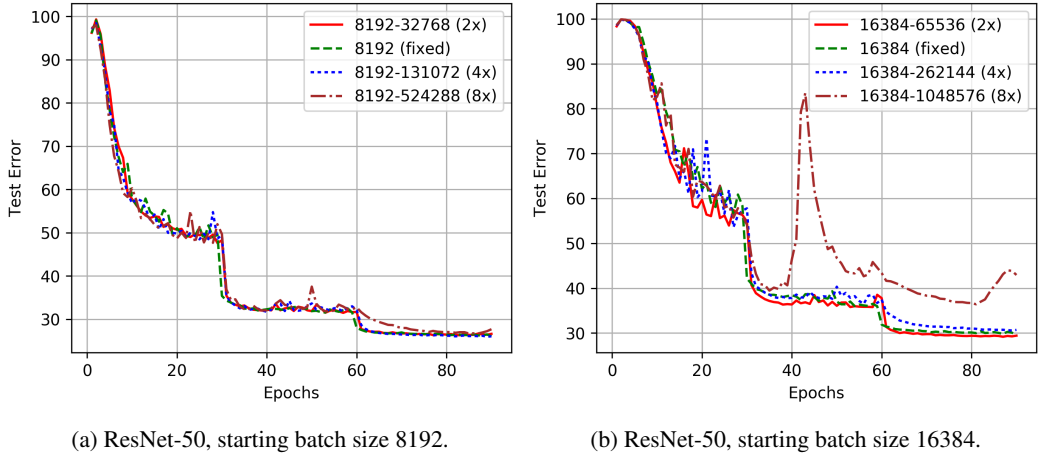

(a) ResNet-50, starting batch size 8192.

(b) ResNet-50, starting batch size 16384.

Figure 7: Comparison of ImageNet test errors curves for adaptive batch sizes with LR warmup and batch size increases of 2x, 4x, and 8x.

Figure 5 shows the evolution of the test error over 90 Epochs of training. We train ResNet-50 with a starting learning rate of $0.1$ which is decayed by $0.1$ every 30 epochs. The network is trained using SGD with momentum of $0.9$ and weight decay of $1 \times 10^{-4}$. For the AdaBatch experiment we double the batch size and decay the learning rate by $0.2$ every 30 epochs. Note that the effective learning rate for AdaBatch and fixed batch sizes is the same for fair comparison. As the results indicate, AdaBatch convergence closely matches the fixed 4096 batch size. Fixed batch sizes of 8192 and 16,384 do not converge to the test errors of fixed batch size 4096 and AdaBatch. These results indicate that AdaBatch attains the same test error as fixed, small batches, but with the advantage of eventually training with a large batch size of 16,384. Due to the progressively larger batch sizes, it is likely that AdaBatch will achieve higher performance and train faster.

Figure 6 illustrates the test errors of large batch size ImageNet training using learning rate warmup over the first $5$ epochs (Goyal et al., 2017). For these experiments we use a baseline batch size of $256$ when linearly scaling the learning rate. All other training parameters remain the same as in Figure 5. Figure 6a compares the test error curves of adaptive batch size 8192-32768 against fixed batch sizes 8192, 16384, and 32768. Figure 6b performs the same experiment but with starting batch sizes of $16384$. In both experiments, we observe that adaptive batch sizes have similar convergence behavior to the small, fixed batch size settings ($8192$ in Fig. 6a and $16384$ in Fig. 6b). Furthermore, the adaptive batch size technique attains lower test errors than large, fixed batch sizes.

Thus far, the adaptive batch size experiments have doubled the batch size at fixed intervals. In Figure 7 we explore the convergence behavior of adaptive batch sizes with increase factors of $2\times$, $4\times$ and $8\times$. We use learning rate warmup with a baseline batch size of 256 for learning rate scaling and with all other training parameters the same as in previous ImageNet experiments. Figure 7a compares the test errors of fixed batch size 8192 and adaptive batch sizes starting at 8192 with batch size increase factors of $2\times$, $4\times$, and $8\times$ and learning rate decayed by 0.2, 0.4, and 0.8, respectively, every 30 epochs. Our results indicate that the test error curves of all adaptive batch size settings closely match the fixed batch size curve. The convergence of adaptive batch size with $8\times$ increase slows after Epoch 60, however the final test error is similar to other curves. This experiment suggests that adaptive batch sizes enable ImageNet training with batch sizes of up to $524,288$ without significantly altering test error. Figure 7b shows test error curves with batch sizes starting at 16384. Unlike Figure 7a, increasing the batch size by $8\times$ results in poor convergence. This is a result of increasing the batch size too much and too early in the training process. Therefore, it is important to tune the batch size increase factor proportional to the starting batch size. With starting batch size of $16384$, we are able to increase the batch size by a factor of $4\times$ without significantly altering final test error and attain a final batch size of $262,144$.

## 5 CONCLUSION

In this paper we have developed an adaptive scheme that dynamically varies the batch size during training. We have shown that using our scheme for AlexNet, ResNet and VGG neural network architectures with CIFAR-10, CIFAR-100 and ImageNet datasets we can maintain the better test accuracy of small batches, while obtaining higher performance often associated with large batches. In particular, our results demonstrate that adaptive batch sizes can attain speedups of up to $6.25\times$ on 4 NVIDIA P100 GPUs with less than $1\%$ accuracy difference when compared to fixed batch size baselines. Also, we have briefly analysed the effects of choosing larger batch size with respect to learning rate, test accuracy and performance as well as work per epoch. Our ImageNet experiments illustrate that batch sizes of up to $524,288$ can be attained without altering test error performance. In the future, we would like to explore the effects of different schedules for adaptively resizing the batch size, including possibly shrinking it to improve convergence properties of the algorithm.

## 6 ACKNOWLEDGEMENTS

The authors would like to thank Cris Cecka, Natalia Gimelshein, Boris Ginsburg, Oleksii Kuchaiev and Sean Treichler for their insightful comments and suggestions.

## A APPENDIX: DETAILED ANALYSIS OF LAYERS

The neural networks used in this work are largely formed from a composition of convolution (Goodfellow et al., 2016), batch normalization (Ioffe & Szegedy, 2015) and fully connected layers (Bishop, 1995), which are briefly reviewed next. During backward propagation, weight updates are always performed using the following rule:

$$W_{new} = W - \frac{\alpha}{r} \Delta W \qquad (8)$$

for some matrix $\Delta W$, batch size $r$ and learning rate $\alpha$.

### A.1 LOSS FUNCTION

The choice of the loss function $\mathcal{L}$ often depends on a particular application. In this paper we will assume, without loss of generality, that we are using cross entropy loss which can be written as

$$\mathcal{L} = \frac{1}{q}\sum_{t=1}^{q}\left(\frac{1}{r}\sum_{s=1}^{r}\mathcal{E}(\mathbf{x})\right) \qquad (9)$$

where index $t$ is over batches, $s$ is over batch elements, and for a particular data sample

$$\mathcal{E}(\mathbf{x}) = -\sum_{i=1}^{M} z_i^* \log(p_i) \qquad (10)$$

with the softmax function

$$p_i = \frac{e^{z_i^{(l)}}}{\sum_{j=1}^{M} e^{z_j^{(l)}}} \qquad (11)$$

and probability $\mathbf{p} = [p_i]$, target $\mathbf{z}^* = [z_i^*]$ as well as computed $\mathbf{z}^{(l)} = [z_i^{(l)}]$ at the output level $l$.

Notice that if the target vectors represents the assignment of inputs into classes and that a given input belongs only to a particular class, then only a single $i$-th component of $\mathbf{z}^*$ is non-zero and therefore

$$\mathcal{E}(\mathbf{x}) = -z_i^* \log(p_i) \qquad (12)$$

### A.2 FULLY CONNECTED LAYER

A fully connected layer with input $\mathbf{x} \in \mathbb{R}^n$, intermediate $\mathbf{y} \in \mathbb{R}^m$ and output $\mathbf{z} \in \mathbb{R}^m$ is written as

$$\mathbf{z} = \mathbf{f}(\mathbf{y}) \qquad (13)$$
$$\mathbf{y} = W\mathbf{x} + \mathbf{b} \qquad (14)$$

where $W \in \mathbb{R}^{m \times n}$ is a matrix of weights, $\mathbf{b} \in \mathbb{R}^m$ is a vector of biases, and $\mathbf{f}(\cdot)$ is a component-wise application of activation function $f(\cdot)$.

In backward propagation, the weight updates are based on the weight gradient components

$$\frac{\partial \mathcal{E}}{\partial w_{ij}} = \left(\frac{\partial \mathcal{E}}{\partial z_i}\right)\left(\frac{\partial z_i}{\partial y_i}\right)\left(\frac{\partial y_i}{\partial w_{ij}}\right) \qquad (15)$$

$$= \left(\frac{\partial \mathcal{E}}{\partial z_i}\right) f'(y_j) x_j \qquad (16)$$

where the first term corresponds to the error gradient component. For the cross entropy loss in (10) at the output layer $l$ the error gradient component can be written as

$$\frac{\partial \mathcal{E}}{\partial z_i} = p_i - z_i^* \qquad (17)$$

while at the hidden layers it can be expressed as

$$\frac{\partial \mathcal{E}}{\partial x_i} = \sum_{j=1}^{n}\left(\frac{\partial \mathcal{E}}{\partial z_j}\right)\left(\frac{\partial z_j}{\partial y_j}\right)\left(\frac{\partial y_j}{\partial x_i}\right) \qquad (18)$$

$$= \sum_{j=1}^{n}\left(\frac{\partial \mathcal{E}}{\partial z_j}\right) f'(y_j) w_{ji} \qquad (19)$$

Therefore, the error gradient in matrix form can be written as

$$\mathbf{v}^{(l)} = (\mathbf{p} - \mathbf{z}^*) \circ \mathbf{f}'(\mathbf{y}^{(l)}) \tag{20}$$

$$\mathbf{v}^{(k)} = \left(W^T \mathbf{v}^{(k+1)}\right) \circ \mathbf{f}'(\mathbf{y}^{(k)}) \tag{21}$$

for $k = 1, ..., l - 1$ and Hadamard (element-wise) matrix product denoted by $\circ$.

Therefore, for a batch of size $r$ the weight update is given by

$$W_{new}^{(k)} = W^{(k)} - \frac{\alpha}{r} \Delta W^{(k)} \tag{22}$$

$$\Delta W^{(k)} = \sum_{i=1}^{r} \mathbf{v}_i^{(k)} \mathbf{x}_i^T \tag{23}$$

where $\alpha$ is the learning rate (Naumov, 2017).

Notice that the computation to perform forward in (13)–(14) and backward propagation in (21)–(23) for the fully connected layer requires $O(mnr)$ flops, which depends linearly on the batch size.

### A.3 CONVOLUTION LAYER

Let $m$ and $n$ be the height and length of the input $A = [a_{gh}] \in \mathbb{R}^{m \times n}$, while $k_1$ and $k_2$ be the dimensions of the kernel $W = [w_{ij}] \in \mathbb{R}^{k_1 \times k_2}$. Also, let $s_1$ and $s_2$ be the vertical and horizontal strides, respectively. Further, let us assume that the input is padded so that divisions are exact when computing the vertical $m' = (m - k_1)/s_1 + 1$ and horizontal $n' = (n - k_2)/s_2 + 1$ dimensions of the output. We denote a 2D convolution $\Theta = [\theta_{gh}] \in \mathbb{R}^{m' \times n'}$ of input $A$ and kernel $W$ as

$$\Theta = A \odot W \tag{24}$$

where

$$\theta_{gh} = \sum_{i=1}^{k_1} \sum_{j=1}^{k_2} w_{ij} a_{k_1+g'-i, k_2+h'-j} \tag{25}$$

$$= \sum_{i=0}^{k_1-1} \sum_{j=0}^{k_2-1} w_{k_1-i, k_2-j} a_{g'+i, h'+j} \tag{26}$$

where $g' = (g - 1)s_1 + 1$ and $h' = (h - 1)s_2 + 1$. Notice that if both strides $s_1 = s_2 = 1$ then $g' = g$ and $h' = h$.

Also, notice that in (25) we index elements of the kernel $W$ in increasing order, while of the input $A$ in decreasing order, and vice-versa in (26). It is important to note that indexing elements of $W$ in increasing order is equivalent to indexing in decreasing order the elements of matrix $P^T W Q$, where permutations $P$ and $Q$ correspond to $180°$ rotation of the convolution kernel $W$.

Then, a 2D convolution layer with output $O = [o_{gh}] \in \mathbb{R}^{m' \times n'}$ can be written as

$$o_{gh} = f(c_{gh}) \tag{27}$$

$$c_{gh} = \theta_{gh} + b_{gh} \tag{28}$$

where $\theta_{gh}$ is defined in (25), $B = [b_{gh}] \in \mathbb{R}^{m' \times n'}$ is a matrix of bias and $f(\cdot)$ is an activation function.

In backward propagation, the weight updates are based on the weight gradient components

$$\frac{\partial \mathcal{E}}{\partial w_{ij}} = \sum_{g=1}^{m'} \sum_{h=1}^{n'} \left(\frac{\partial \mathcal{E}}{\partial o_{gh}}\right) \left(\frac{\partial o_{gh}}{\partial c_{gh}}\right) \left(\frac{\partial c_{gh}}{\partial w_{ij}}\right) \tag{29}$$

$$= \sum_{g=1}^{m'} \sum_{h=1}^{n'} \left(\frac{\partial \mathcal{E}}{\partial o_{gh}}\right) f'(c_{gh}) a_{k_1-i+g', k_2-j+h'} \tag{30}$$

where the first term corresponds to the error gradient component, which can be expressed as

$$\frac{\partial \mathcal{E}}{\partial a_{st}} = \sum_{g=1}^{m'} \sum_{h=1}^{n'} \left(\frac{\partial \mathcal{E}}{\partial o_{gh}}\right) \left(\frac{\partial o_{gh}}{\partial c_{gh}}\right) \left(\frac{\partial c_{gh}}{\partial a_{st}}\right) \quad (31)$$

$$= \sum_{g=1}^{m'} \sum_{h=1}^{n'} \left(\frac{\partial \mathcal{E}}{\partial o_{gh}}\right) f'(c_{gh}) w_{k_1-s+g',k_2-t+h'} \quad (32)$$

$$= \sum_{i=1}^{k'_1} \sum_{j=1}^{k'_2} \left(\frac{\partial \mathcal{E}}{\partial o_{s'+i,t'+j}}\right) f'(c_{s'+i,t'+j}) w_{i',j'} \quad (33)$$

where we have used an index switch $i' = (i-1)s_1+1+(s-1)\%s_1$, $j' = (j-1)s_2+1+(t-1)\%s_2$, $k'_1 = \lfloor(k_1-1)/s_1\rfloor + 1$, $k'_2 = \lfloor(k_2-1)/s_2\rfloor + 1$, $s' = \lfloor(s-k_1)/s_1\rfloor$ and $t' = \lfloor(t-k_2)/s_2\rfloor$ to obtain (33). Also, notice that $s-k_1+1 \leq g' \leq s$ and $t-k_2+1 \leq h' \leq t$ in (32), while $s'+i > 0$ and $t'+j > 0$ in (33), otherwise the elements with corresponding indices in the summation are considered to be 0. Finally, % denotes a modulo operation.

Therefore, the error gradient in matrix form can be written as

$$V^{(k)} = \left(V^{(k+1)} \odot (\bar{P}^T W \bar{Q})\right) \circ \mathbf{f}'(C^{(k)}) \quad (34)$$

for $k = 1, ..., l-1$ and where permutation and selection matrices $\bar{P}$ and $\bar{Q}$ correspond to selection of elements with strides $s_1$ and $s_2$ as well as $180°$ rotation of the convolution kernel $W$ [1].

Therefore, for a batch of size $r$ the weight update is given by

$$W_{new}^{(k)} = W^{(k)} - \frac{\alpha}{r} \Delta W^{(k)} \quad (35)$$

$$\Delta W^{(k)} = \sum_{i=1}^{r} (P^T A Q) \odot V^{(k)} \quad (36)$$

where permutations $P$ and $Q$ correspond to $180°$ rotation of the input $A$.

Notice that for the convolutional layer the computation to perform forward propagation in (27)–(28) requires $O(k_1 k_2 m' n' r)$ flops and backward propagation in (34)–(36) requires $O(k'_1 k'_2 m n r)$ flops, which both depend linearly on the batch size.

---

[1] For example, let a matrix

$$W = \begin{pmatrix} w_{11} & w_{12} & w_{13} & w_{14} \\ w_{21} & w_{22} & w_{23} & w_{24} \\ w_{31} & w_{32} & w_{33} & w_{34} \\ w_{41} & w_{42} & w_{43} & w_{44} \\ w_{51} & w_{52} & w_{53} & w_{54} \\ w_{61} & w_{62} & w_{63} & w_{64} \end{pmatrix}$$

Notice that we can select elements with stride $s_1 = 3$ and $s_2 = 2$, as well as perform $180°$ rotation of the result

$$\bar{P}^T W \bar{Q} = \begin{pmatrix} w_{43} & w_{41} \\ w_{13} & w_{11} \end{pmatrix}$$

using permutation and selection matrices

$$\bar{P}^T = \begin{pmatrix} 0 & 0 & 0 & 1 & 0 & 0 \\ 1 & 0 & 0 & 0 & 0 & 0 \end{pmatrix} \text{ and } \bar{Q} = \begin{pmatrix} 0 & 1 \\ 0 & 0 \\ 1 & 0 \\ 0 & 0 \end{pmatrix}$$

A.4 BATCH NORMALIZATION LAYER

A batch normalization layer works on the entire batch. Let the batched input to this layer be $X = [\mathbf{x}_1, ..., \mathbf{x}_r] \in \mathbb{R}^{m \times r}$ and output be $Z = [\mathbf{z}_1, ..., \mathbf{z}_r] \in \mathbb{R}^{m \times r}$, then it can be written as

$$Y = \frac{1}{\sqrt{r}} X \left( I - \frac{1}{r} \mathbf{e}\mathbf{e}^T \right) \tag{37}$$

$$D = \text{diag}(\mathbf{f}([\sum_{j=1}^{r} y_{1j}^2, ..., \sum_{j=1}^{r} y_{mj}^2])) \tag{38}$$

$$\hat{X} = \sqrt{r} D^{-1} Y \tag{39}$$

$$Z = W\hat{X} + \mathbf{b}\mathbf{e}^T \tag{40}$$

where $W$ is a diagonal matrix of weights, $\mathbf{b}$ is a vector of bias, $\mathbf{f}(\cdot)$ is a component-wise application of $f(x) = \sqrt{x + \epsilon}$ and $\mathbf{e}^T = [1, ..., 1]$. Notice that vector $\frac{1}{r} X \mathbf{e}$ and diagonal matrix $D$ correspond to mean $\mu_\mathcal{B}$ and variance $\sigma_\mathcal{B}^2$ for each row of the batched input, respectively (Ioffe & Szegedy, 2015).

In backward propagation, the weight updates are based on the weight gradient components

$$\frac{\partial \mathcal{E}}{\partial w_{ii}} = \sum_{j=1}^{r} \left( \frac{\partial \mathcal{E}}{\partial z_{ij}} \right) \left( \frac{\partial z_{ij}}{\partial w_{ii}} \right) \tag{41}$$

$$= \sum_{j=1}^{r} \left( \frac{\partial \mathcal{E}}{\partial z_{ij}} \right) \hat{x}_{ij} \tag{42}$$

where the first term corresponds to the error gradient component, which can be expressed as

$$\frac{\partial \mathcal{E}}{\partial x_{ij}} = \sum_{k=1}^{r} \left( \frac{\partial \mathcal{E}}{\partial z_{ik}} \right) \left( \frac{\partial z_{ik}}{\partial \hat{x}_{ik}} \right) \left( \frac{\partial \hat{x}_{ik}}{\partial x_{ij}} \right) \tag{43}$$

$$= \sum_{k=1}^{r} \left( \frac{\partial \mathcal{E}}{\partial z_{ik}} \right) w_{ii} \frac{\sqrt{r}}{d_{ii}^2} \left( d_{ii} \frac{\partial y_{ik}}{\partial x_{ij}} - y_{ik} \frac{\partial d_{ii}}{\partial x_{ij}} \right) \tag{44}$$

$$= \sum_{k=1}^{r} \left( \frac{\partial \mathcal{E}}{\partial z_{ik}} \right) \frac{w_{ii}}{d_{ii}} \left( (\delta_{kj} - \frac{1}{r}) - \frac{y_{ik}}{d_{ii}^2} \sum_{h=1}^{r} y_{ih} (\delta_{hj} - \frac{1}{r}) \right) \tag{45}$$

where $\delta_{kj}$ is the Kronecker delta.

Therefore, the error gradient in matrix form can be written as

$$\hat{Y} = Y(I - \frac{1}{r}\mathbf{e}\mathbf{e}^T) \tag{46}$$

$$\hat{V}^{(k+1)} = V^{(k+1)}(I - \frac{1}{r}\mathbf{e}\mathbf{e}^T) \tag{47}$$

$$\hat{U}^{(k+1)} = (V^{(k+1)} \circ Y)\mathbf{e}\mathbf{e}^T \tag{48}$$

$$V^{(k)} = D^{-1} W \left( \hat{V}^{(k+1)} - D^{-2}(\hat{U}^{(k+1)} \circ \hat{Y}) \right) \tag{49}$$

for $k = 1, ..., l-1$ and Hadamard (element-wise) matrix product denoted by $\circ$.

Therefore, for a batch of size $r$ the weight update is given by

$$W_{new}^{(k)} = W^{(k)} - \frac{\alpha}{r} \Delta W^{(k)} \tag{50}$$

$$\Delta W^{(k)} = \text{diag}([\sum_{j=1}^{r} v_{1j}^{(k)} \hat{x}_{1j}, ..., \sum_{j=1}^{r} v_{mj}^{(k)} \hat{x}_{mj}]) \tag{51}$$

Notice that the computation to perform forward in (37)–(40) and backward propagation in (46)–(51) for the batch normalization layer requires $O(mr)$ flops, which depends linearly on the batch size.